\documentclass[letterpaper]{article} 
\usepackage{aaai2026}  
\usepackage{times}  
\usepackage{helvet}  
\usepackage{courier}  
\usepackage[hyphens]{url}  
\usepackage{graphicx} 
\usepackage{natbib}  
\usepackage{caption} 
\usepackage{algorithm}
\usepackage{algorithmic}
\usepackage{booktabs}
\usepackage{graphicx}
\usepackage{amsmath}
\usepackage{amssymb}
\usepackage{booktabs}
\usepackage{mathrsfs}
\usepackage{graphicx}
\usepackage{subcaption}
\usepackage{enumitem}
\usepackage{url}

\usepackage[table]{xcolor}
\usepackage{newfloat}
\usepackage{listings}
\DeclareCaptionStyle{ruled}{labelfont=normalfont,labelsep=colon,strut=off} 
\floatstyle{ruled}
\newfloat{listing}{tb}{lst}{}
\floatname{listing}{Listing}
\title{ICL-Router: In-Context Learned Model Representations for LLM Routing}
\author{
    Chenxu Wang\textsuperscript{\rm 1,2}\thanks{This work was done during his internship at Shanghai Artificial Intelligence Laboratory.} \equalcontrib\quad
    Hao Li\textsuperscript{\rm 2,3}\footnotemark[1]\equalcontrib  \quad
    Yiqun Zhang\textsuperscript{\rm 2,4}\footnotemark[1] \quad
    Linyao Chen\textsuperscript{\rm 2,5}\footnotemark[1]  \quad\\
    Jianhao Chen\textsuperscript{\rm 2,6}\footnotemark[1]  \quad
    Ping Jian\textsuperscript{\rm 7}  \quad
    Peng Ye\textsuperscript{\rm 2}  \quad
    Qiaosheng Zhang\textsuperscript{\rm 2}  \quad
    Shuyue Hu\textsuperscript{\rm 2}\thanks{Corresponding Author}
}

\affiliations{
    \textsuperscript{\rm 1}Fudan University \quad
    \textsuperscript{\rm 2}Shanghai Artificial Intelligence Laboratory \quad \\
    \textsuperscript{\rm 3}Northwestern Polytechnical University \quad
    \textsuperscript{\rm 4}Northeastern University \quad \\
    \textsuperscript{\rm 5}The University of Tokyo \quad
    \textsuperscript{\rm 6}Nanjing University \quad
    \textsuperscript{\rm 7}Beijing Institute of Technology \\
    \texttt{\{wangchenxu,zhangqiaosheng,hushuyue\}@pjlab.org.cn} \quad
    \texttt{li.hao@mail.nwpu.edu.cn}

}

\usepackage{bibentry}

\nocopyright

\begin{document}

\maketitle

\begin{abstract}
Large language models (LLMs) often exhibit complementary strengths. Model routing harnesses these strengths by dynamically directing each query to the most suitable model, given a candidate model pool. However, routing performance relies on accurate model representations, and adding new models typically requires retraining, limiting scalability.
To address these challenges, we propose a novel routing method using in-context vectors to represent model capabilities. The method proceeds in two stages. First, queries are embedded and projected into vectors, with a projector and LLM-based router trained to reconstruct the original queries, aligning vector representations with the router’s semantic space. Second, each candidate model is profiled on a query set, and the router learns---based on in-context vectors of query and model performance---to predict whether each model can correctly answer new queries. Extensive experiments demonstrate that our method achieves state-of-the-art routing performance in both in-distribution and out-of-distribution tasks. Moreover, our method allows for seamless integration of new models without retraining the router. The code is available at \url{https://github.com/lalalamdbf/ICL-Router}.

\end{abstract}


\section{Introduction}

Large language models (LLMs) have demonstrated exceptional capabilities performance across various tasks, including mathematical reasoning~\cite{acemath2024, chervonyi2025goldmedalistperformancesolvingolympiad}, code generation~\cite{deepseekai2024deepseekcoderv2breakingbarrierclosedsource,wang2025coevolvingllmcoderunit}, logical understanding~\cite{liu2025logicofthoughtinjectinglogiccontexts, wang2025thought}, and STEM problem-solving~\cite{wu2025sharp,ma2025generalreasoneradvancingllmreasoning}. Despite these advances, \emph{no} single model consistently outperforms others in every domain. Instead, models often exhibit complementary strengths and weaknesses, shaped by differences in their training data, architecture, and optimization. 

An emerging line of research explores \emph{routing} methods. The intuition is straightforward: if models excel in different tasks, then dynamically assigning the most capable model in a model pool to each query might maximize overall performance~\cite{chen2024routerdc,lu-etal-2024-routing,zhuang2024embedllm,zhang2025capability}.
However, in practice, achieving this goal is \emph{far} from trivial. A key challenge is that \emph{effective routing critically depends on an accurate understanding of each model’s capabilities}. For example, given two general-purpose models, it is difficult to tell which tasks LLaMA~\cite{grattafiori2024llama} excels at versus those where Gemma~\cite{team2025gemma} performs better, unless both are \emph{extensively} evaluated across a wide range of tasks. 
This challenge is further exacerbated by the rapid pace of LLM development. As new models are frequently released from time to time, routing methods must be able to \emph{incrementally} adapt to these new models. Otherwise, the cost of repeating large-scale evaluations and retraining routers will quickly become prohibitive, ultimately limiting the ability of routing methods to benefit from model scaling.

Although some recent efforts have begun to address this challenge, 
existing approaches fall short of addressing scalability and evaluation efficiency. For instance, RouterDC~\cite{chen2024routerdc} uses dual contrastive learning to jointly train query and model embeddings, aligning each query with its optimal model.
Similarly, EmbedLLM~\cite{zhuang2024embedllm} applies binary cross-entropy loss to train an embedding-based router that predicts query–model compatibility. 
However, both RouterDC and EmbedLLM operate on a fixed set of LLMs; incorporating a new model requires retraining the router.
More recently, MODEL-SAT~\cite{zhang2025capability} represents models using a hand-crafted ``capability instruction'' based on performance on the MMLU benchmark~\cite{hendryckstest2021}. This removes the need to retrain the router when adding new models, but requires manual instruction design for other benchmarks and prior knowledge of the capabilities they measure.

\begin{figure*}[t]
\centering
\includegraphics[width=.96\textwidth]{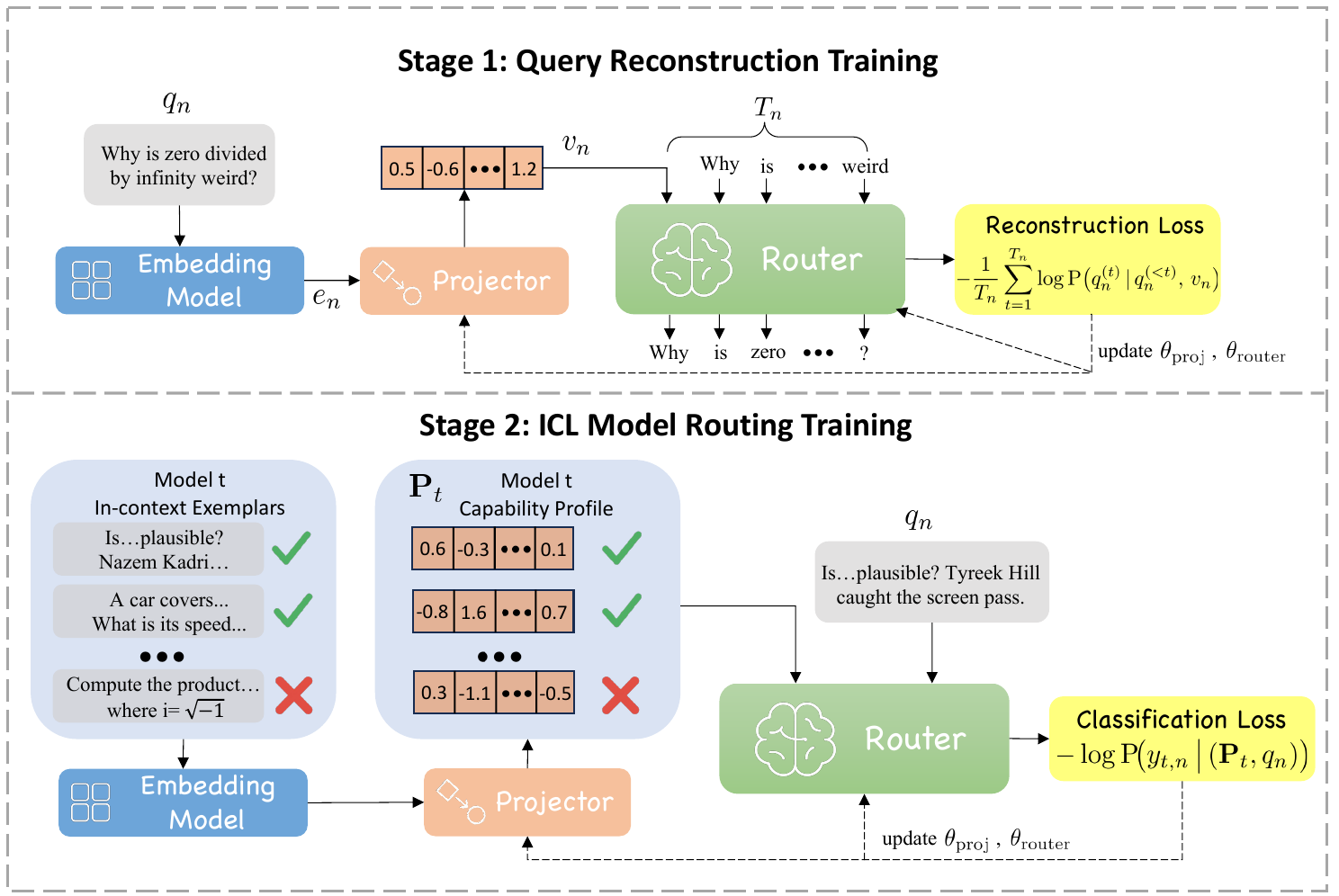} 
\caption{The two-stage ICL-Router framework. (1) Query Reconstruction Training: The projector is trained to align the embedding model and router dimensions, while the router reconstructs queries from projected vectors to learn their semantics. (2) ICL Model Routing Training: Each model’s capabilities are encoded as in-context vectors, and the router is trained to predict whether a given model can handle a specific query.}
\label{icl_framework}
\end{figure*}

To this end, in this paper, we propose \textbf{ICL-Router}, a novel method that leverages in-context learned representations to characterize model capabilities. We hypothesize that a model’s performance on diverse queries can serve as in-context exemplars, as the way a model responds to various queries naturally reflects its unique capability profile.
However, including hundreds or thousands of queries directly as in-context information would result in an unmanageably long context window. Inspired by the recent concept of \emph{in-context vectors}~\cite{zhuang2024vector,liu2024context}, we instead transform these exemplars into compact in-context vectors, substantially reducing context length.
At inference time, the router uses these vectors---compact representations of each model’s capability profile---to estimate the probability that each model can correctly answer a given query. When a new LLM is introduced, it is quickly evaluated on a small set of queries to generate in-context vectors (i.e. its capability profile), which is then fed to the router, enabling immediate inclusion without retraining.

More specifically, our ICL-Router consists of three main components: an embedding model, a projector, and an LLM-based router. It is trained in two stages, as shown in Figure \ref{icl_framework}. In the first stage, queries are embedded and projected into vector representations; the projector and the LLM-based router are co-trained to align their semantic spaces, allowing the router to interpret the information encoded in the vectors effectively.
In the second stage, each candidate LLM in the model pool is evaluated on a small set of queries. The resulting performance is paired with the vector representations of the queries to form the model’s capability profile. Using these query-performance pairs as in-context vectors, the router is then trained to predict whether a given model can successfully handle new queries.

We evaluate the performance of ICL-Router on 10 widely used benchmarks (AGIEval, AIME, BBH, HumanEval, KORBench, LogicBench, MBPP, MMLU-CF, MMLUPro, and OlympiadBench) across both in-distribution and out-of-distribution scenarios. Experimental results show that ICL-Router achieves state-of-the-art (SOTA) routing performance.
Specifically, on in-distribution tasks, ICL-Router outperforms the best-performing single LLM in the model pool (Deepseek-R1-distill-Qwen-7B) by 7.2 absolute points (Table 1);  it also surpasses RouterDC by 3.9 points, EmbedLLM by 2.2 points, and MODEL-SAT by 4.6 points.
Moreover, unlike EmbedLLM and RouterDC---which assume a fixed model pool and require retraining to incorporate new models---the ICL-Router seamlessly benefits from model scaling. As more models are added, it significantly outperforms MODEL-SAT (Figures 2 and 3), demonstrating superior scalability and adaptability. Beyond model scalability, the ICL-Router also benefits from incorporating a modest number of in-context exemplars, which further enhances routing performance as shown in Figures~\ref{fig:id_icl_quantities} and~\ref{fig:ood_icl_quantities}.

Our key contributions are summarized as follows:

\begin{itemize}

\item We propose characterizing model capabilities through vector representations of query-performance pairs, offering a new perspective on model representation.

\item We propose a novel routing method, ICL-Router, that decouples model capability profiling from routing, enabling scalable integration of new models without retraining.

\item Empirical results demonstrate that ICL-Router achieves SOTA routing performance across multiple domains in both in-distribution and OOD tasks. 

\end{itemize}

\section{Related Work}

\textbf{In-Context Learning.} In-context learning (ICL) enables LLMs to perform new tasks using only demonstrations provided in the prompt. This capability was first effectively demonstrated by GPT-3~\cite{brown2020language}, which achieved strong few-shot and even zero-shot performance across diverse tasks by learning from in-context examples. With the advent of LLMs supporting longer context windows, in-context learning has shown remarkable improvements, especially when provided with hundreds or even thousands of examples~\cite{li2023incontextlearningdemonstrationexamples, chen-etal-2023-many,bertsch2025context}. Nevertheless, this leads to a substantial increase in both the context window size and the cost of inference. To overcome these limitations, recent research has explored the use of in-context vectors to enable LLMs to process context information more efficiently and flexibly. For example,~\citet{zhuang2024vector} propose Vector-ICL, which projects continuous input data from various modalities into the model’s embedding space, allowing LLMs to learn directly from vectorized demonstrations in the prompt. Meanwhile,~\citet{liu2024context} focus on improving context selection and demonstration efficiency, further reducing inference costs while maintaining high performance across a range of in-context learning scenarios. Building on these advances, our work explores the use of in-context vectors as model representations for model routing.  

\noindent\textbf{Model Routing.} Model routing aims to efficiently allocates queries to the optimal model without invoking all available models. Recent research in this area generally falls into two categories: studies that focus on balancing performance with computational cost~\cite{ong2024routellm,feng2024graphrouter,wang2025mixllm,jitkrittum2025universal,zhang2025router}, and those that prioritize maximizing model performance~\cite{llm-blender-2023,lu-etal-2024-routing,chen2024routerdc, zhuang2024embedllm, zhang2025capability,zhang2025avengers}. For the former,  GraphRouter~\cite{feng2024graphrouter} utilizes a graph-based approach to allocate queries among models, reducing computational overhead while maintaining performance. MixLLM~\cite{wang2025mixllm} further optimizes the trade-off between resource consumption and accuracy by dynamically selecting models based on input complexity and cost-effectiveness. Router-R1~\cite{zhang2025capability} treats routing as a sequential decision process, embodying the router itself as an LLM that alternates between “think” and “route” steps to progressively decompose tasks, invoke models dynamically, and jointly balance performance and cost. For the latter, ZOOTER~\cite{lu-etal-2024-routing} introduces a reward-guided approach to query routing, utilizing tag-based label enhancement to promote greater training stability. RouterDC~\cite{chen2024routerdc} introduces a dual-contrastive learning approach to better align queries and model representations. EmbedLLM~\cite{zhuang2024embedllm} proposes an encoder–decoder framework that  leverages compact model embeddings and query embeddings to predict routing accuracy. MODEL-SAT~\cite{zhang2025capability} employs aptitude-based instruction tuning to characterize model capabilities and dynamically route instructions to the most suitable LLMs. However, existing performance-oriented routing approaches encounter two major challenges: the model representations they construct are generally overly simplistic, and incorporating new models is inefficient, usually requiring substantial retraining and large-scale inference. To overcome these limitations, the proposed ICL-Router employs a two-stage training process to learn semantically rich model representations and enables the efficient integration of new models using only a small set of representative queries.

\section{ICL-Router}
\subsection{Overview}
We introduce ICL-Router, a framework for constructing in-context learned representations to support LLM routing. It comprises three main components: an embedding model, a projector, and an LLM-based router.
Consider a set of candidate LLMs. 
At inference time, each query is first encoded using the embedding model and projector to obtain its vector representation. The router then takes this query vector along with the in-context vectors representing each candidate LLM’s capabilities as input, and selects the model best suited to handle the query.
As illustrated in Figure \ref{icl_framework}, our approach features a two-stage training strategy: (1)  \textbf{Query Reconstruction Training}, which aims to enable the projector align the dimension between the embedding model and the router, while also allowing the router to interpret the vectors generated by the embedding model. (2) \textbf{ICL Model Routing Training}, which utilizes the in-context model performance vectors as input to train the router to select the most suitable LLM for each query. 

\subsection{Query Reconstruction Training}
\label{icl_stage1}
The core objective of this stage is to co-train the projector and the router. The projector learns to map vectors from an embedding model into the routing model's semantic space, while the router simultaneously learns to understand and interpret these incoming vector representations.

Given a set of queries $\mathcal{Q} = \{q_n \mid n = 1, \ldots, N\}$, each query $q_n$ is first encoded into an embedding by the embedding model $f_{\mathrm{emb}}$. The embedding $e_n$ corresponding to query $q_n$ is defined as:
\begin{equation}
e_n = f_{\mathrm{emb}}(q_n) \in \mathbb{R}^{d_\mathrm{Emb}}.
\end{equation}

Next, each embedding $e_n$ is projected into a vector $v_n$ by a projector $f_{\mathrm{proj}}$, which is trained to align the embedding dimension with the LLM-based router's input space:
\begin{equation}
v_n = f_{\mathrm{proj}}(e_n) \in \mathbb{R}^{d_\mathrm{Router}}.
\end{equation}

The router is trained to reconstruct the original query $q_n$ from its projected vector $v_n$ in an autoregressive manner. Specifically, at each step, it maximizes the conditional probability of the current token $q_n^{(t)}$ given the previously generated tokens $q_n^{(<t)}$ and the projected vector $v_n$. The training objective is thus defined by minimizing the reconstruction loss:
\begin{equation}
\mathcal{L}_{\mathrm{rec}}(\theta_{\mathrm{proj}},\theta_{\mathrm{router}})
=
-\frac{1}{NT_n}\sum_{n=1}^{N}
\sum_{t=1}^{T_n}
\log
\mathrm{P}\bigl(q_n^{(t)}\,|\,q_n^{(<t)},\, v_n\bigr),
\end{equation}
where $q_n^{(t)}$ denotes the $t$-th token, $q_n^{(<t)}$ represents all preceding tokens, and $T_n$ is the number of tokens. By jointly training the projector and router to minimize this reconstruction loss, we encourage the projection module to produce vectors aligned with the router’s semantic space, ensuring that projected queries are interpretable by the router for downstream routing.

\subsection{ICL Model Routing Training}
\label{icl_stage2}

Building upon the vector representation learned during query reconstruction, this stage trains the router to assign each query to the most appropriate LLM. By leveraging in-context vectors that represent each LLM's capabilities as input, the router learns to match queries with the LLM that is most likely to provide correct answers.

Consider a set of LLMs, denoted as  $ \mathbb{M} = \{\mathcal{M}_t : t = 1, \ldots, T \}$, and a set of queries $\mathscr{Q} = \{q_k \mid k = 1, \ldots, K\}$. For each model $\mathcal{M}_t$, we construct a capability profile $\mathbf{P}_t$ by evaluating the model’s responses to each query in $\mathscr{Q}$. The query set $\mathscr{Q}$ consists of challenging queries, specifically those that only a few LLMs are able to answer correctly. Formally, for each model $\mathcal{M}_t$, we define its capability profile $\mathbf{P}_t$ as:
\begin{equation}
\mathbf{P}_t = ((v_1, c_1), (v_2, c_2), \ldots, (v_K, c_K)),
\end{equation}
where $v_k$ is the vector representation of each query $q_k\in\mathscr{Q}$, which is encoded through the embedding model and the projector, and $c_k$ indicates whether $\mathcal{M}_t$ answered query $q_k$ correctly, with ‘Yes’ and ‘No’ denoting correct and incorrect responses, respectively.

These capability profiles provide a detailed characterization of model capabilities. 
We jointly train the projector and the router to predict whether a model can accurately handle a query, conditioned on its capability profile. 
Formally, the projector and the router are optimized together using a cross-entropy loss:
\begin{equation}
\mathcal{L}_{\mathrm{ce}}(\theta_{\mathrm{proj}},\theta_{\mathrm{router}})
=
- \frac{1}{TN} \sum_{t=1}^T\sum_{n=1}^N
\log
\mathrm{P}
\!\bigl(
y_{t,n}\,\bigl|\,
(\mathbf{P}_t, q_n)
\bigr),
\end{equation}
where $T$ is the number of LLMs, $N$ is the number of queries, and $y_{t,n}$ is the ground-truth label indicating whether model $\mathcal{M}_t$ answers query $q_n$ correctly.
Conditioning on these vectors allows the router to reason about the relative strengths and weaknesses of each candidate model, without requiring an excessively long context window.

\subsection{Inference and Scalable Model Incorporation}
During inference, for each new query $q'$, the router combines it with each model’s capability profile $\mathbf{P}_t$ and outputs the probability that model $\mathcal{M}_t$ will answer correctly. The model with the highest predicted probability is selected:
\begin{equation}
\mathcal{M}^{*}
\;=\;
\underset{t=1,\dots,T}{\arg\max}
\;p\bigl(\text{‘Yes’}\mid \mathcal{M}_{t}, q'\bigr),
\end{equation}
where $p\bigl(\text{‘Yes’}\mid \mathcal{M}_{t}, q'\bigr)$ denotes the router-estimated probability that $\mathcal{M}_t$ will produce a correct answer for the query.
When a new model $\mathcal{M}_{T+1}$ is introduced, we simply evaluate it on the same query set $\mathscr{Q}$ to construct its capability profile $\mathbf{P}_{T+1}$ with the embedding model and the projector. The capability profile of this new model can then be incorporated into the routing process without any additional retraining. Therefore, our approach allows for rapid, plug-and-play integration of new LLMs in dynamic environments.

\begin{table*}[t]
  \centering
  \setlength{\tabcolsep}{5pt}
  \label{tab:8b_models}
  \begin{tabular}{lccccccc}
    \toprule
    \textbf{Method} & \textbf{OlympiadBench} & \textbf{BBH}  & \textbf{LogicBench} & \textbf{MMLUPro}  & \textbf{MBPP} & \textbf{Avg.} \\
    \midrule
    DeepSeek-R1-Distill-Qwen-7B         & 66.24 & 72.87  & 78.03 & 58.84  & 69.52 & 69.10 \\
    Llama-3.1-8B-Instruct               & 16.93  & 59.47 & 69.47 & 49.38   & 60.03 & 51.06 \\
    Llama-3.1-Nemotron-Nano-8B-v1       & 74.26  & 53.24  & 65.43 & 50.40  & 79.21 & 64.51\\
    Qwen2.5-7B-Instruct                 & 38.02 & 71.54  & 71.93 & 57.49  & 71.99 & 62.19\\
    cogito-v1-preview-llama-8B          & 16.58 & 75.62 & 68.43 & 58.04  & 60.82 & 55.90 \\
    Gemma-2-9B-IT                      & 13.71  & 62.19 & 68.27 & 55.42  & 60.31 & 51.98 \\
    Internlm3-8B-Instruct               & 33.12 & 68.64 & 72.83  & 56.49  & 56.88 & 57.59 \\
    GLM-4-9B-Chat                       & 16.88 & 51.14 & 69.53 & 48.58   & 61.10 & 49.65 \\
    \midrule
    Random Router & 34.47 & 64.34  & 71.41   & 54.33    & 64.98  & 57.91   \\
    LLM Router & 46.03   & 69.54   & 73.19   & 55.87      & 66.55   & 62.36   \\
    Max Expert & \textbf{74.26}  & 75.62  & 78.03  & 58.84    & 79.21  & 73.19  \\
    RouterDC  & 73.56 & 73.49 & 77.24 & 58.20  & \underline{79.11} & 72.32 \\
    EmbedLLM  & 71.45  & \underline{79.02} & \underline{78.92} & \underline{64.06} & 77.34 & \underline{74.16} \\
    MODEL-SAT  & 73.02   & 71.14   & 74.80    & 63.61   & 76.00   & 71.71   \\
    \textbf{ICL-Router (ours)} & \underline{74.16}   & \textbf{80.52}    & \textbf{79.03}     & \textbf{67.53}   & \textbf{80.53}   & \textbf{76.30}   \\
    \bottomrule
  \end{tabular}
  \caption{Comparison of ICL-Router with baselines on in-distribution tasks. The best is in \textbf{bold} and the second-best is \underline{underlined}.}
  \label{tab1:in_domain}
\end{table*}

\begin{table*}[t]
  \centering
  \setlength{\tabcolsep}{5pt}
  \label{tab:8b_models}
  \begin{tabular}{lcccccc}
    \toprule
    \textbf{Method} & \textbf{AIME}  & \textbf{KORBench} & \textbf{MMLU-CF}  & \textbf{AGIEval} & \textbf{HumanEval}  &\textbf{Avg.} \\
    \midrule
    DeepSeek-R1-Distill-Qwen-7B         & 45.50 & 50.30  & 56.70 & 62.56 & 83.78 & 59.77  \\
    Llama-3.1-8B-Instruct               & 3.67  & 43.67  & 62.16 & 54.40 & 65.43  & 45.87  \\
    Llama-3.1-Nemotron-Nano-8B-v1       & 58.00  & 30.91  & 49.76 & 47.64 & 91.10 & 55.48  \\
    Qwen2.5-7B-Instruct                 & 8.00 & 38.31  & 64.98 & 62.23 & 84.27 & 57.58 \\
    cogito-v1-preview-llama-8B          & 2.67 & 44.99 & 62.11 & 58.56 & 67.80 & 50.83   \\
    Gemma-2-9B-IT                      & 10.00  & 37.16 & 64.33 & 60.60 & 63.23 & 51.82   \\ 
    Internlm3-8B-Instruct               & 5.83 & 38.53 & 64.58  & 64.01 & 67.98 & 52.99   \\
    GLM-4-9B-Chat                       & 1.67 & 35.67  & 60.76 & 59.11 & 67.68  & 50.18  \\
    \midrule
    Random Router & 15.79 & 37.80  & 60.67  & 59.02 & 73.91   & 49.44   \\
    LLM Router & 30.33   & 40.67   & 60.48  & 58.08  & 75.20   & 52.95    \\
    Max Expert & \textbf{58.00}  & \underline{50.30}  & 64.98  & 64.01 & \textbf{91.10}   & \underline{65.68}     \\
    RouterDC  & \textbf{58.00} & 46.44 & 59.50 & 61.49  & \underline{90.20} & 63.13 \\
    EmbedLLM  & 53.72 & 46.90 & \textbf{65.22} & 64.99 & 83.29 & 62.82  \\
    MODEL-SAT  & \underline{56.84}   & 44.53  & 64.45  & 60.08  & 89.02  & 62.99    \\
    \textbf{ICL-Router (ours)} & \textbf{58.00}   & \textbf{53.03}   & \underline{64.99}   & \textbf{67.29}  & 89.04   & \textbf{66.47}     \\
    \bottomrule
  \end{tabular}
  \caption{Comparison of ICL-Router with baselines on OOD tasks. The best is in \textbf{bold} and the second-best is \underline{underlined}.}
   \label{tab2:ood}
\end{table*}

\section{Experiments}

\subsection{Datasets}

We evaluate our method on 10 benchmarks, including OlympiadBench~\cite{chervonyi2025goldmedalistperformancesolvingolympiad}, AIME, MBPP~\cite{austin2021programsynthesislargelanguage}, HumanEval~\cite{zhong2023agievalhumancentricbenchmarkevaluating}, BBH~\cite{suzgun2022challengingbigbenchtaskschainofthought}, LogicBench~\cite{parmar2024logicbenchsystematicevaluationlogical}, KORBench~\cite{ma2025korbenchbenchmarkinglanguagemodels}, MMLUPro~\cite{wang2024mmluprorobustchallengingmultitask}, AGIEval~\cite{zhong2023agievalhumancentricbenchmarkevaluating}, and MMLU-CF~\cite{zhao2024mmlucfcontaminationfreemultitasklanguage}. To evaluate the generalization of ICL-Router, we partition these benchmarks into in-distribution and out-of-distribution (OOD) sets. We designate AIME, HumanEval, KORBench, AGIEval and MMLU-CF as OOD datasets, reserving them exclusively for testing. Referring to the settings in RouterDC~\cite{chen2024routerdc}, the other benchmarks are considered in-distribution and split into training and test sets at a 7:3 ratio. Detailed information about these datasets can be found in Appendix A.1.

\subsection{Implementation Settings}

Our framework is composed of three main components: an embedding model, a projector, and an LLM-based router. We adopt \textit{Qwen3-Embedding-8B}~\cite{qwen3embedding} as the embedding model; notably, this model is used solely to generate query vectors and does not participate in the training process. The projector is implemented as a two-layer multi-layer perceptron (MLP) that facilitates the transformation between the embedding space and the router. For the router, we employ Qwen2.5-7B-Instruct, given its good performance and versatility. For the LLM pool, we consider eight widely-used open-source LLMs.

Our training process consists of two stages. In the first stage, we begin by training only the projector for one epoch with a learning rate of 2e-5. This is followed by two epochs of joint training, where the router is introduced with a learning rate of 5e-6. we construct a representative query set by sampling 500 challenging queries that only a few LLMs are capable of answering correctly. We jointly train the projector and router for five epochs with learning rates of 1e-5 and 2e-6, respectively. The training is repeated three times using different random seeds. For both stages, we set the batch size to 32. To ensure stable results during evaluation, we sample each routed LLM 10 times, setting the temperature to 0.3 and top-$p$ to 1.0, and then report the average accuracy.

\subsection{Baselines}

We compare \textbf{ICL-Router} with the following baselines:

\begin{itemize}

\item \textbf{Random Router}: Selects a model at random from the pool of candidate LLMs for each query.

\item \textbf{LLM Router}: A prompt-based approach that leverages an LLM (Qwen2.5-7B-Instruct) to choose models based on natural-language descriptions of their performance profiles.

\item \textbf{Max Expert}: Serves as a strong baseline by selecting the best-performing model for each dataset.

\item \textbf{RouterDC}~\cite{chen2024routerdc}: Trains query and model embeddings using dual contrastive learning, pulling queries closer to suitable models and clustering semantically similar queries within the representation space.

\item \textbf{EmbedLLM}~\cite{zhuang2024embedllm}: An encoder–decoder framework that employs compact model and query embeddings to predict model-query compatibility, with the router trained via binary cross-entropy loss.

\item \textbf{MODEL-SAT}~\cite{zhang2025capability}: Leverages capability instruction tuning by converting candidate model performance into textual descriptions, which are embedded and passed to a trainable LLM that dynamically predicts the most suitable model for each query.
\end{itemize}

For more detailed implementation specifics, please refer to Appendix A.2.

\begin{figure}[t]
  \centering
  \includegraphics[width=\columnwidth]{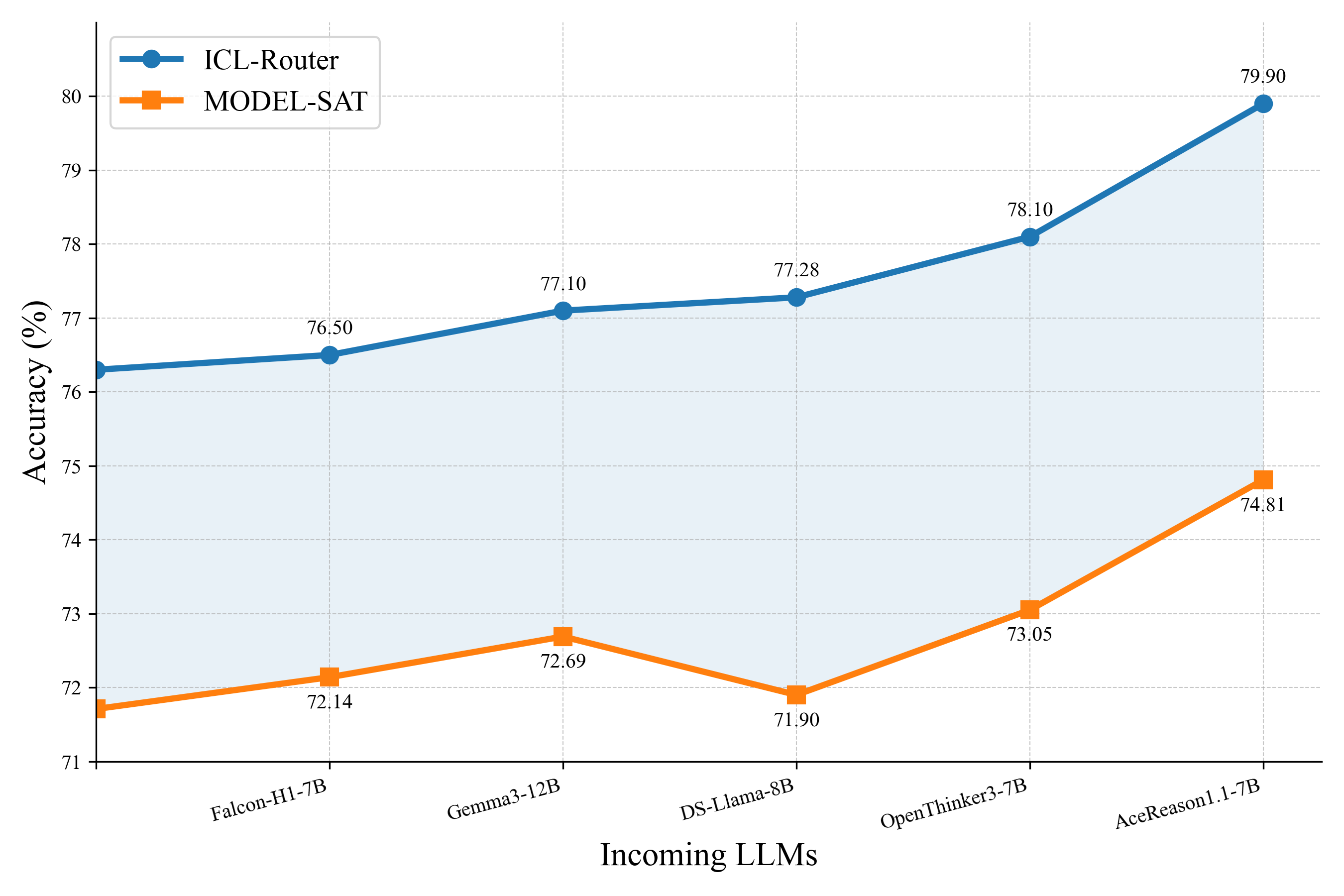} 
  \caption{Effects of integrating new LLMs on in-distribution routing performance.}
  \label{fig:id}
\end{figure}

\subsection{Main Results}
Table \ref{tab1:in_domain} presents the performance comparison across five in-distribution datasets. \textbf{ICL-Router achieves the highest average accuracy of 76.30\%, outperforming RouterDC by 4.08\%, EmbedLLM by 2.14\% and MODEL-SAT by 4.59\%.} Notably, ICL-Router attains the best results on 4 out of the 5 datasets, demonstrating its consistent superiority across a diverse range of tasks. Furthermore, ICL-Router surpasses the Max Expert—which selects the best-performing model for each dataset—by an average of 3.11\%. This improvement is primarily driven by substantial accuracy gains on several tasks: MMLUPro (+8.69\%), BBH (+4.9\%), LogicBench (+1\%), and MBPP (+1.32\%). These results indicate that ICL-Router can match each query to the most suitable model based on its specific characteristics, rather than relying solely on domain-level or task-based selection.

Table \ref{tab2:ood} shows the performance comparison across five OOD tasks. Remarkably, \textbf{ICL-Router achieves an average accuracy of 66.47\%, surpassing RouterDC by 3.34\%, EmbedLLM by 3.65\% and MODEL-SAT by 3.48\%.} This result highlights ICL-Router’s robust performance across diverse OOD scenarios, confirming its effectiveness beyond in-distribution settings. Moreover, ICL-Router significantly outperforms the Max Expert baseline on KORBench (+2.73\%) and AGIEval (+3.28\%), while achieving nearly identical results on AIME, MMLU-CF, and HumanEval. This demonstrates that ICL-Router not only generalizes well across OOD tasks, but also reliably matches or surpasses the performance of the strongest individual model on each benchmark.

Overall, the results demonstrate that ICL-Router consistently outperforms all baseline methods on both in-distribution and OOD tasks. This underscores the strength of our approach in introducing accurate and semantically rich model representations, which effectively capture subtle differences in the capabilities of candidate models and enable more precise routing. The robust and stable performance observed across benchmarks further validates the effectiveness and generalization ability of our method.

\begin{figure}[t]
  \centering
  \includegraphics[width=\columnwidth]{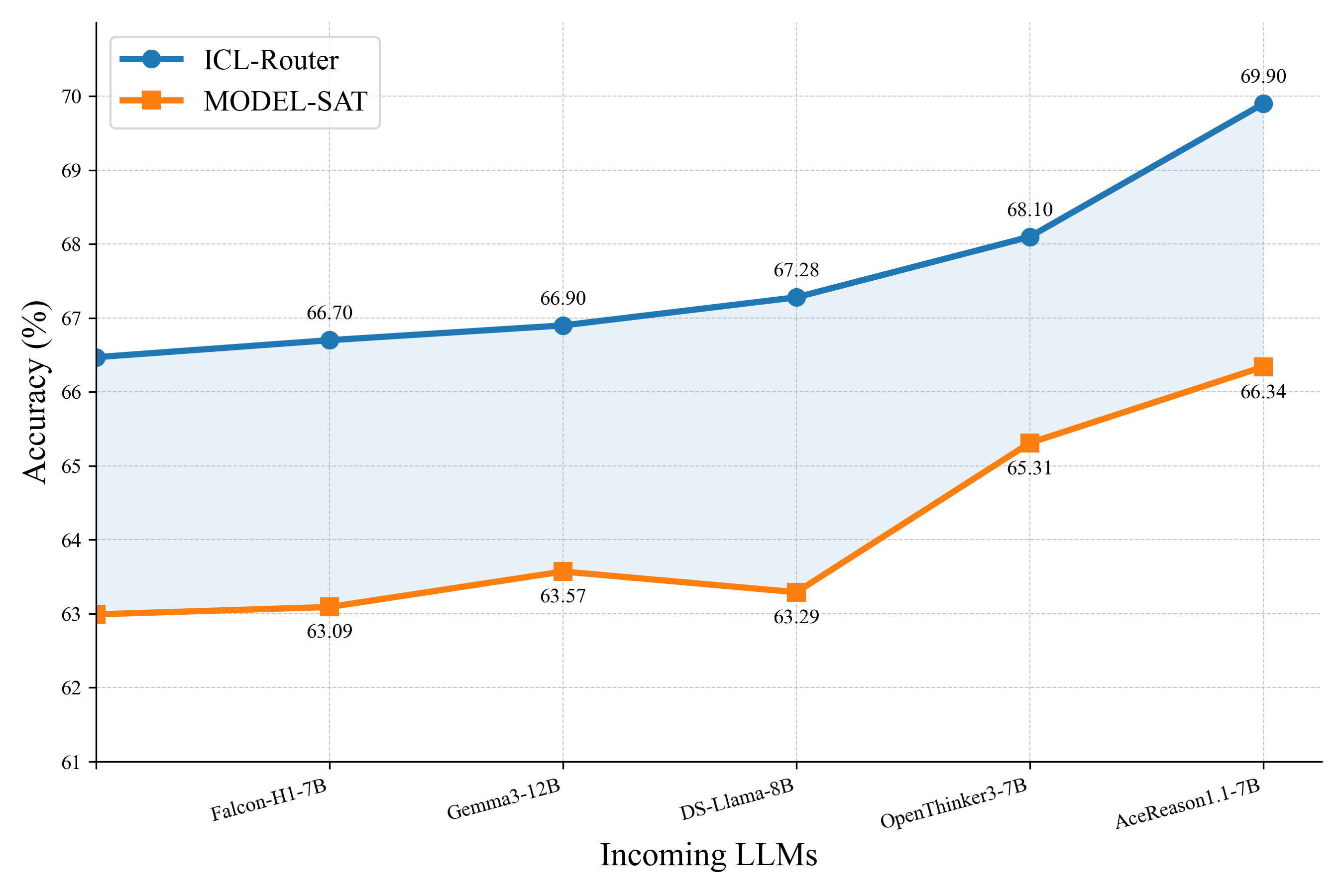}  
  \caption{Effects of integrating new LLMs on out-of-distribution (OOD) routing performance.}
  \label{fig:ood}
\end{figure}

\subsection{Scalability to New Models}

We conduct experiments to assess the scalability of our method on a set of recently released and competitive LLMs, including Falcon-H1-7B-Instruct~\cite{falconh1}, Gemma3-12B-IT~\cite{team2025gemma}, DeepSeek-R1-Llama-8B~\cite{guo2025deepseek}, OpenThinker3-7B~\cite{guha2025openthoughtsdatarecipesreasoning}, and AceReason-Nemotron-1.1-7B~\cite{liu2025acereason}. As illustrated in Figures \ref{fig:id} and \ref{fig:ood}, ICL-Router demonstrates a consistent upward trajectory in accuracy for both in-distribution and out-of-distribution (OOD) scenarios. For example, in the in-distribution setting, ICL-Router’s accuracy improves from 76.3\% to 79.9\% as more models are integrated; a similar trend is observed on OOD tasks, where performance rises from 66.4\% to 69.9\%. In contrast, MODEL-SAT exhibits less stable gains, with accuracy improvements that are inconsistent and sometimes fluctuate as additional models are introduced.

Importantly, at higher accuracy levels, where further improvements are typically more difficult, \textbf{ICL-Router consistently maintains and even expands its performance lead as more models are added}. This steady and reliable progress demonstrates that our method effectively leverages the strengths of an expanding model pool. These results show that our approach remains robust and dependable, even as new models are continually integrated in real-world settings.

\begin{table}[t]
\centering
\begin{tabular}{lccc}
\toprule
\textbf{Embed.} & \textbf{Para.}/\textbf{Dim.} & \textbf{ID} & \textbf{OOD} \\
\midrule
mxbai-embed-large-v1             & 0.3B/1024  & 74.45  & 64.55    \\
bge-m3   & 0.6B/1024      & 75.13  &  65.14   \\
stella-en-1.5B-v5         & 1.5B/2048    & 75.51  & 65.45   \\
gte-Qwen2-7B-instruct  & 7B/3584      & \underline{76.03}  & \underline{66.06}   \\
Qwen3-8B-Embedding         & 8B/4096      & \textbf{76.30}  & \textbf{66.47}    \\
\bottomrule
\end{tabular}
\caption{Embedding model comparison in terms of parameter size and embedding dimension. Performance is evaluated on both in-distribution (ID) and out-of-distribution (OOD) scenarios.}
\label{fig:emb}
\end{table}

\subsection{Analysis}

To better understand the effectiveness of ICL-Router. we analyze how its performance is influenced by three key elements — the embedding model, the number of in-context exemplars, and the query reconstruction training stage.

\noindent\textbf{Embedding Model.} The choice of embedding model plays a pivotal role in determining the quality of vector representations used for downstream routing. Specially, we evaluate the impact of various embedding models on both in-distribution and OOD accuracy by comparing five embedding models that differ in parameter size and embedding dimensions, including \textit{
mxbai-embed-large-v1}~\cite{emb2024mxbai} , \textit{bge-m3}~\cite{bge-m3}, \textit{stella-en-1.5B-v5}~\cite{zhang2025jasperstelladistillationsota}, \textit{gte-Qwen2-7B-instruct}~\cite{li2023towards}, \textit{Qwen3-8B-Embedding}~\cite{qwen3embedding}. As shown in Table~\ref{fig:emb}, a clear trend emerges: as model size increases—from the smallest, \textit{mxbai-embed-large-v1} (0.3B/1024), to the largest, \textit{Qwen3-8B-Embedding} (8B/4096)—in-distribution accuracy improves from 74.45\% to 76.30\%, while out-of-distribution (OOD) accuracy rises from 64.55\% to 66.47\%. This stepwise improvement across all five evaluated models indicates that \textbf{stronger embedding models consistently produce more robust and generalizable vector representations}, leading to better downstream routing performance.

\begin{figure}[t]
  \centering
  \includegraphics[width=\columnwidth]{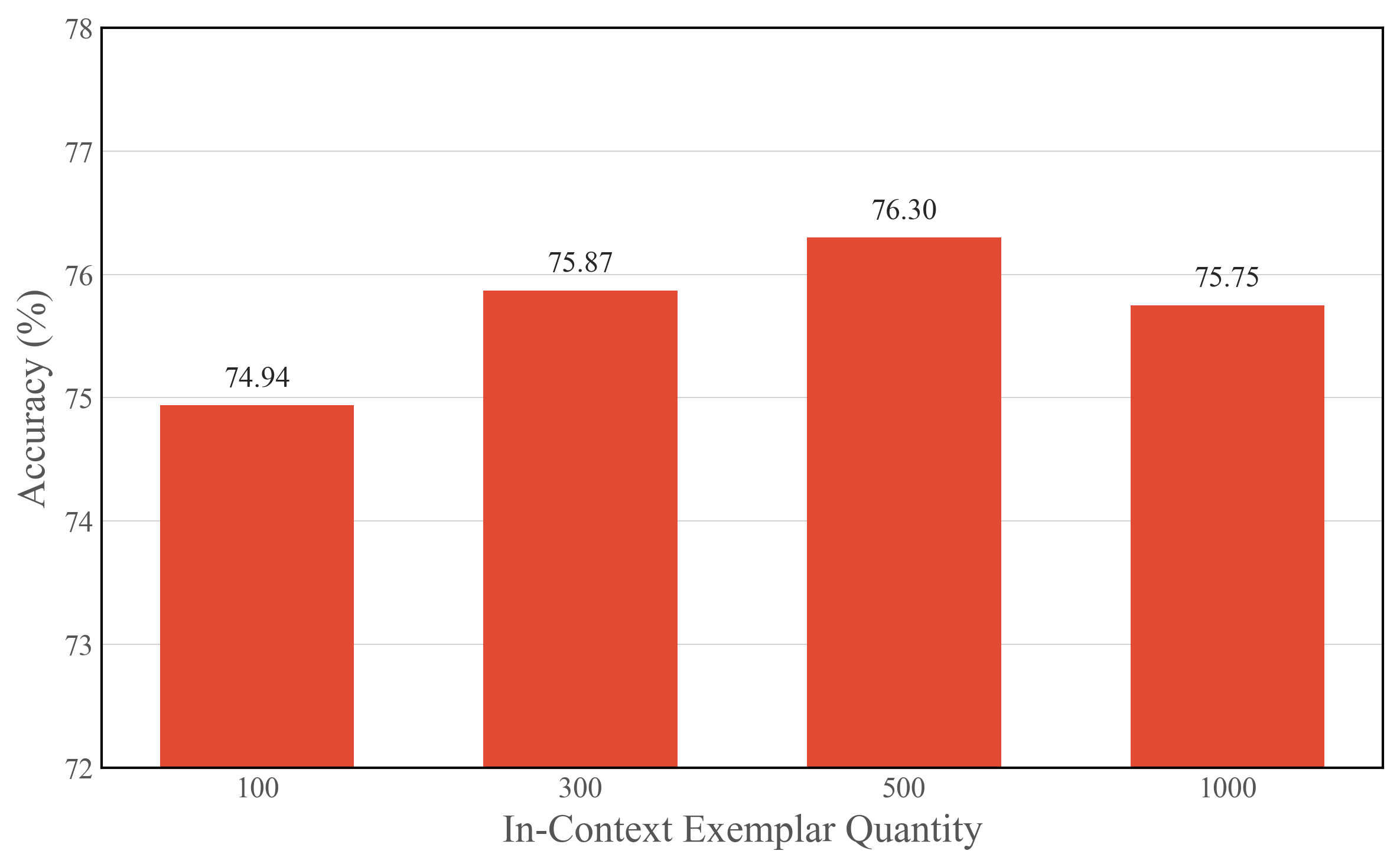}  
  \caption{Effects of in-Context exemplar quantity on in-distribution performance.}
  \label{fig:id_icl_quantities}
\end{figure}

\begin{figure}[t]
  \centering
  \includegraphics[width=\columnwidth]{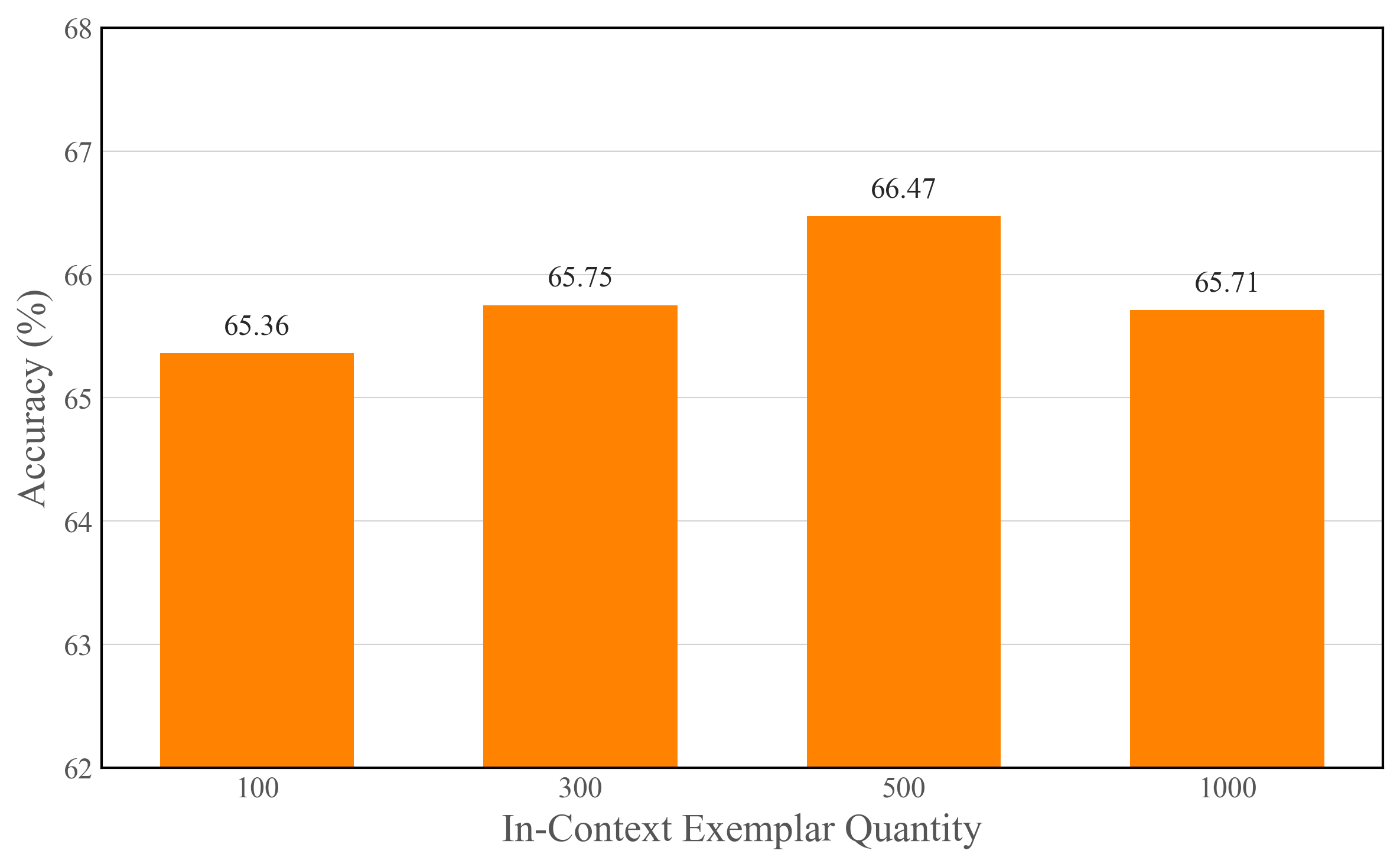}  
  \caption{Effects of in-context exemplar quantity on OOD performance.}
  \label{fig:ood_icl_quantities}
\end{figure}

\noindent\textbf{Effects of In-Context Exemplar Quantity.} To further investigate the impact of the number of in-context exemplars on routing performance, we conduct ablation experiments by varying the quantity of exemplars used to construct each model’s capability profile. Specifically, we assess performance with 100, 300, 500, and 1000 in-context exemplars, analyzing both in-distribution and out-of-distribution (OOD) scenarios. As shown in Figures \ref{fig:id_icl_quantities} and \ref{fig:ood_icl_quantities}, increasing the number of in-context exemplars consistently improves both in-distribution and out-of-distribution routing accuracy up to a certain point. For in-distribution tasks, accuracy increases from 74.94\% with 100 exemplars to a peak of 76.30\% at 500 exemplars, before slightly dropping to 75.75\% at 1000 exemplars. A similar trend is observed for out-of-distribution performance, where accuracy rises from 65.36\% (100 exemplars) to 66.47\% (500 exemplars), but then plateaus or slightly declines to 65.71\% with 1000 exemplars.  These results suggest that while \textbf{adding more exemplars generally enriches model capability profiles and enhances routing, there is a point of diminishing returns—likely due to redundancy or increased noise as the exemplar set grows too large}. This observation is consistent with findings from recent studies on many-shot in-context learning \cite{agarwal2024many,li2024long,zhang2025more}, which report that including too many context examples can sometimes degrade performance rather than improve it. Overall, using a moderate number of exemplars strikes the best balance between informativeness and efficiency, supporting robust and generalizable routing performance.

\begin{table}[t]
\centering
\renewcommand{\arraystretch}{1.15}
\setlength{\tabcolsep}{14pt}
\begin{tabular}{lcc}
\toprule
\textbf{Method} & \textbf{ID} & \textbf{OOD} \\
\midrule
w/o QRT   & 74.01  & 64.06   \\

\textbf{ICL-Router}    & \textbf{76.30}  & \textbf{66.47}    \\
\bottomrule
\end{tabular}
\caption{Ablation study of query reconstruction training (QRT) on in-distribution (ID) and OOD scenarios.}
\label{tab:ab_qrt}
\end{table}

\noindent\textbf{Effects of Query Reconstruction Training.} Table~\ref{tab:ab_qrt} demonstrates that \textbf{removing query reconstruction training (QRT) leads to a clear decline in performance on both in-distribution and OOD tasks, with accuracy dropping by 2.29\% and 2.41\%}, respectively. This result highlights the importance of QRT for aligning the embedding and router spaces, enabling the router to more effectively leverage query representations during downstream routing. Without this training stage, the router is less able to capture and differentiate the relevant semantics required for accurate model selection.

\section{Conclusion}
In this paper, we propose a new insight on model representation for LLM routing by characterizing model capabilities through in-context vectors. By leveraging compact vector-based profiles that summarize how each LLM performs on challenging queries, our approach constructs semantically rich model representations that capture the diverse capabilities of LLMs. ICL-Router decouples capability profiling from routing, allowing plug-and-play integration of new models without retraining the router. Empirical results on 10 benchmarks demonstrate that ICL-Router achieves state-of-the-art performance across both in-distribution and OOD settings, demonstrating the effectiveness and scalability of in-context learned model representations for LLM routing.

\bibliography{aaai2026}

\appendix
\setcounter{secnumdepth}{2}
\setcounter{table}{0}
\makeatletter
\renewcommand\thesubsection{\thesection.\arabic{subsection}}
\section{Technical Appendices}

\subsection{Datasets}

We evaluate our method on ten widely used benchmarks, as summarized in Table~\ref{tab:datasets}. Detailed descriptions for each dataset are provided below.

\begin{itemize}[leftmargin=9pt]

\item \textbf{OlympiadBench}: A challenging benchmark derived from problems featured in international mathematics and physics Olympiads. In our settings, we focus on 674 mathematics questions presented in plain text, excluding those that require diagrams or images.

\item \textbf{BBH}: A challenging benchmark derived from Big-Bench Hard (BBH), consisting of 23 difficult tasks designed to evaluate advanced reasoning abilities of large language models. In our setting, we select 1,080 examples from the original dataset, covering diverse domains that require complex, multi-step reasoning.

\item \textbf{LogicBench}: A natural language QA dataset specifically designed to systematically evaluate the logical reasoning capabilities of large language models, covering 25 distinct inference-rule reasoning patterns from propositional, first-order, and non-monotonic logics. The dataset includes both binary (yes/no) and multiple-choice QA tasks, with each question focused on a single inference rule to enable precise measurement of accuracy. For our experiments, we selected a subset consisting of 1,000 examples from the original dataset.

\item \textbf{MMLUPro}: A large-scale academic and professional QA benchmark expressly built to probe the breadth and depth of knowledge reasoning in cutting-edge language models, MMLU-Pro spans 14 disciplines and 57 sub-fields, pairing each conceptually demanding prompt with ten answer options to lower chance accuracy and amplify discriminative power. Every item is presented in multiple-choice form only, forcing models to engage in fine-grained recall, elimination, and cross-domain reasoning under consistent conditions. For our experiments we employed a stratified subset of 1,500 questions selected from the full collection.

\item \textbf{MBPP}: Designed to test code generation skills, MBPP is a collection of 974 simple Python programming tasks, each described in everyday language and paired with a starter function and hidden test cases. Models are evaluated by their ability to turn these natural-language prompts into working code that passes all tests, focusing on true programming understanding rather than memorization.

\item \textbf{AIME}: An Olympiad-level math benchmark based on 60 problems from the 2024 and 2025 \textit{American Invitational Mathematics Examination}. The dataset features numeric-answer questions spanning algebra, combinatorics, geometry, and number theory.

\item \textbf{KORBench}: A deliberately knowledge-orthogonal reasoning benchmark, KORBench forges 25 entirely novel rules across five domains—Operation, Logic, Cipher, Puzzle, and Counterfactual—and couples each rule with ten problems, creating conceptually unfamiliar challenges that force models to infer the governing pattern instead of leaning on memorized facts. Every problem is cast in a ten-option multiple-choice format to curb guessing and sharpen discriminative evaluation. For our study, we used a curated subset of 1,250 question–answer items drawn from the full suite.

\item \textbf{MMLU-CF}: A contamination-free variant of the MMLU benchmark, MMLU-CF consists of 10,000 carefully filtered multiple-choice questions spanning a wide range of academic and professional subjects, each with four answer options to ensure robust evaluation of reasoning ability. All questions are presented in multiple-choice format, requiring models to demonstrate genuine subject understanding and cross-domain reasoning. For our experiments, we selected a subset of 1,000 questions from the full dataset.

\item \textbf{AGIEval}: A human-centric benchmark consists of 20 tasks sourced from high-quality standardized exams covering diverse academic and professional subjects. The dataset rigorously evaluates models on understanding, reasoning, knowledge recall, and calculation abilities. For our experiments, we selected a subset of 1,576 examples from the full dataset.

\item \textbf{HumanEval}: A code generation benchmark introduced by OpenAI, HumanEval consists of 164 hand-written Python programming problems, each including a function signature, a docstring prompt, and unit tests for automatic evaluation.

\end{itemize}

\begin{table}[t]
\centering

\begin{tabular}{lll}
\toprule
\textbf{Dataset} & \textbf{Metrics} & \textbf{Size} \\
\midrule
OlympiadBench       & Accuracy, 0-shot &   674 \\
BBH            & Accuracy, 3-shot &  1,080 \\
LogicBench      & Accuracy, 0-shot &  1,000 \\
MMLUPro             & Accuracy, 0-shot & 1,500 \\
MBPP                & Pass@1, 0-shot   & 974   \\
AIME                & Accuracy, 0-shot &   60 \\
KORBench            & Accuracy, 3-shot & 1,250 \\
MMLU-CF          & Accuracy, 0-shot & 1,000 \\
AGIEval           & Accuracy, 0-shot & 1,576 \\
HumanEval           & Pass@1, 0-shot   & 164   \\

\bottomrule
\end{tabular}
\caption{Detailed information of the datasets.}
\label{tab:datasets}
\end{table}

\subsection{Baselines}

\textbf{LLM Router}: In our approach, Qwen2.5-7B-Instruct acts as a router. It processes the incoming query together with model profiles and selects the most appropriate model to handle each query. This routing process is entirely training-free, depending exclusively on the natural language profiles constructed for each model. To automate the creation of these profiles, we first evaluate all candidate models on the training dataset to obtain their performance metrics. We then combine these metrics with task descriptions and prompt GPT-4.1 to generate the model profiles automatically. For both the profile generation and routing inference stages, we maintain consistent settings by fixing the temperature to 0.6 and top-$p$ to 1.0.
\vspace{0.5em}

\noindent\textbf{RouterDC}: To enable a fair comparison with our approach, we adapt the official implementation by substituting the encoder with \textit{Qwen3-8B-Embedding}. We employ DeepSpeed for distributed training across eight NVIDIA A800-80G GPUs, setting the per-GPU batch size to 4 and maintaining all other hyperparameters as in the original setup.
\vspace{0.5em}

\noindent\textbf{EmbedLLM}: We follow the official implementation, replacing the query encoder with \textit{Qwen3-8B-Embedding} to allow for a fair comparison with our approach, and adjust the input layer dimensions as needed. To enhance training stability, we raise the batch size to 32,768, while keeping all other hyperparameters consistent with the original settings.
\vspace{0.5em}

\noindent\textbf{MODEL-SAT}: Since the official implementation is incomplete, we re-implement the primary method, reproduce the codebase, and report the results. We use \textit{Qwen3-8B-Embedding} as the embedding model and Qwen2.5-7B-Instruct as the router, connecting them via a two-layer MLP projector. For efficient training, we leverage DeepSpeed for distributed multi-GPU training across eight NVIDIA A800-80GB GPUs, with a batch size of 4 per GPU. The learning rates are set to 1e-6 for the embedding model, 2e-6 for the router, and 5e-5 for the projector. Initially, we fine-tune only the projector for approximately 1,000 steps; afterward, we continue fine-tuning all model parameters for the remainder of the training. A warmup ratio of 0.1 is applied to stabilize the early training stage.

\subsection{Challenging Query Set Construction}
To construct the set of 500 challenging queries, we selected 125 queries each where exactly 1, 2, 3, or 4 models (given a pool of 8 models) produced the correct answer. The intuition is that all-correct or all-wrong queries fail to differentiate model capability. Note that this set belongs to the full training set of every baseline method we compared. That is, no additional data was introduced for our method.

\subsection{Cost Evaluation}
Although we use a 7B-scale model as the router, it generates only a small number of tokens during inference (e.g., 8 tokens for 8 candidate models). In contrast to the routed model, which typically produce hundreds or even thousands of tokens, the routing overhead in our approach remains well within an acceptable range.

\subsection{Limitations}

Due to limitations in computational resources and the substantial time required to collect data for larger-scale models, our current study primarily focuses on the model pool consisting of small-parameter LLMs. Additionally, the benchmarks utilized in this study are designed for general evaluation and do not specifically assess the chat or instruction-following capabilities of LLMs. We leave the exploration of these scenarios for future work.

\end{document}